\documentclass[letterpaper, 10 pt, conference]{ieeeconf}  

\IEEEoverridecommandlockouts                              

\usepackage{times} 
\usepackage{amsmath} 
\usepackage{amssymb}  
\usepackage{algorithm}
\usepackage{algorithmic}

\usepackage{graphicx}
\usepackage{booktabs}
\usepackage{color}

\usepackage{url}
\usepackage{multirow}
\usepackage{subfigure}
\usepackage{import}
\usepackage{transparent}
\usepackage{algorithm}
\usepackage{epsfig} 
\usepackage{xcolor}
\usepackage{pdfcolmk}
\usepackage{amstext}
\usepackage{ulem}
\usepackage{tikz}

 \usepackage[linewidth=1pt]{mdframed}
\usepackage{soul} 
\usepackage{graphicx} 
\usepackage{float} 
\usepackage{subfigure} 
\usepackage{caption}
\usepackage{multirow}
 
\usepackage{ulem}
\usepackage{makecell}

\title{\LARGE \bf
Uncovering the Secrets of Human-Like Movement: A Fresh Perspective on Motion Planning
}

\author{ Lei Shi$^{*}$, Qichao Liu$^{*}$, Cheng Zhou$^{\dag}$, Wentao Gao, Haotian Wu, Yu Zheng, Xiong Li
\thanks{* denotes co-first author}
\thanks{\dag \ denotes corresponding author
}
\thanks{
The authors are with Tencent Robotics X, Shenzhen, Guangdong, China. {\tt\scriptsize \{chowchzhou\} @tencent.com}}%
\thanks{
Lei Shi is also with University of Wisconsin-Madison,
        Madison, WI, US.{\tt\scriptsize \{lshi222\} @wisc.edu}.  \ 
        }%
\thanks{
Qichao Liu is also with University of of Wisconsin-Madison,
        Madison, WI, US.{\tt\scriptsize \{qliu426\} @wisc.edu}.  \ 
        }%
\thanks{Wentao Gao is with Faculty of STEM, University of South Australia, Adelaide, Australia.
        {\tt\scriptsize \{gaowy014\} @mymail.unisa.edu.au}}%
\thanks{Lei Shi, Wentao Gao conducted this work during their internship at Tencent Robotics X.
        }%
}
\begin{document}

\captionsetup{font={small}}

\maketitle
\thispagestyle{empty}
\pagestyle{empty}

\begin{abstract}

\textcolor{black}{This article aims to uncover the secrets of human-like movement from a fresh perspective on motion planning. For the voluntary movement of the human body, we analyze the coordinated movement mechanism and the compliant movement mechanism of the human body from the perspective of human biomechanics. Based on human motion mechanisms, we propose an optimal control framework that integrates compliant control dynamics, optimizing robotic arm motion through the solution of a response time matrix. This matrix determines the timing parameters for joint movements, transforming the system into a time-parameterized optimal control problem. The model focuses on the interaction between active and passive joints under external disturbances, enhancing the system’s adaptability and compliance. This method not only achieves optimal trajectory generation but also strikes a dynamic balance between precision and compliance. Experimental results on both a manipulator and a humanoid robot validate the effectiveness of this approach.
 }

\end{abstract}

\section{Introduction}

Humanoid research seeks to equip robots with human-like bodies, enabling agile locomotion~\cite{radosavovic2024real}, dexterous manipulation~\cite{zhou2023differential}, and reasoning intelligence~\cite{haarnoja2024learning}. However, most existing studies focus on replicating human motion data in robots, assuming shared physical configurations. These data-driven approaches often overlook the fundamental force transmission and coordination mechanisms of human movement, limiting robotic performance in complex tasks requiring precise manipulation and coordination~\cite{ijspeert2013dynamical,osa2018algorithmic}. Few studies explore how to transfer the evolved human motion mechanisms into robots~\cite{razavian2023body,krotov2022motor}, which could enhance robots' capability for more advanced movement and manipulation.

%
%
%
\begin{figure*}[htbp]
\captionsetup{font={footnotesize}}
\centering %
\includegraphics[width=0.7\textwidth]{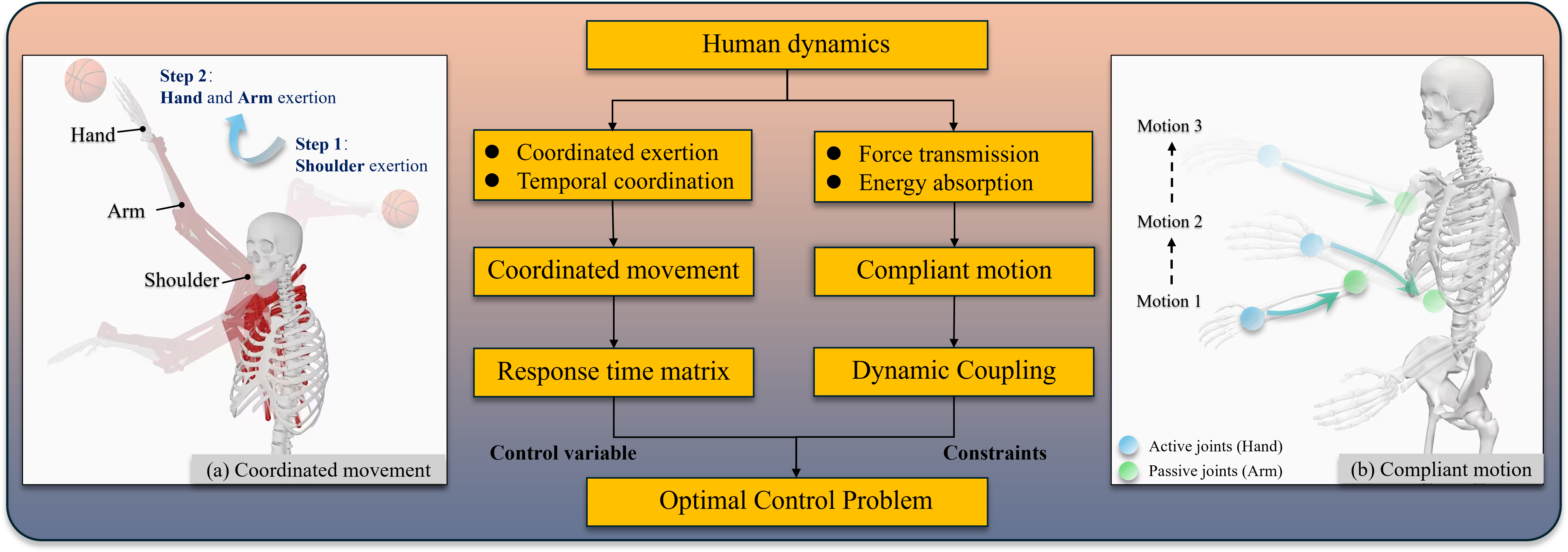}
\caption{Human-Like Motion
Process: (a) Coordinated Movement: Focuses on the torque transmission and temporal synchronization between multiple joints, ensuring movement coordination and efficiency. (b) Compliant Motion: Considers the dynamic behavior of passive joints during force transmission from active joints, including force propagation and energy absorption characteristics, ensuring compliant response and stability under external disturbances.}
\vspace{-0.5cm}
\end{figure*}
Human movement~\cite{ugawa2020voluntary,virameteekul2022we} can generally be divided into three categories: reflex movement, voluntary movement, and rhythmic movement. 
%
%
%
%
%
%
%
%
%
%
%
%
%
%
%
%
{Furthermore, the  motion of human being is not always synchronous. The time of movement can transmit information such as perceived confidence, naturalness, and even the perceived object weight, etc~\cite{zhou2017expressive,venture2019robot}.}
The motion form of the human body is a typical dynamic system, and we can study it from the perspective of optimal control.
%
%
Numerical solution methods of optimal control are solved off-line including direct method and indirect method, collocation-based algorithm and shooting-based algorithm. While effective for some applications, these methods are not well-suited for human-like motion control in robots with high degrees of freedom~\cite{schaal2007dynamics}.
%
%
%
%
%
%


{To summarize, the contributions of this paper include}: 
\begin{itemize}
\item Inspired by the analysis of human body movement, this paper extracts two key characteristics—coordinated and compliant movement;
\item This paper analyzes human coordinated movement and proposes a dynamic model using a response time matrix and an inertia-damping-spring system to integrate coordination and compliance into robotic motion control.
\item This paper presents a hierarchical control framework, where the lower level couples the dynamics of active and passive joints to achieve flexible buffering between joints, while the upper level incorporates a response time matrix for precise coordination control.
\end{itemize}


\section{Related Work}
\label{sec:II}

Dynamic Movement Primitives (DMP) 
is a method for trajectory imitation learning~\cite{ijspeert2013dynamical}.
%
Besides, the nonlinear dynamic systems can also be modelled as Gaussian mixture model(GMM) from a set of demonstrations~\cite{khansari2011learning}. And this can also be extended to estimate non-linear dynamics of objects~\cite{khansari2011learning}.
The neural circuit that generates rhythmic motor activity is called the central pattern generator (CPG).
%
Related theories and algorithms are also applied to biomimetic robotic fish to realize bionic movement~\cite{wang2019control}.
The energy analysis of a CPG-controlled robotic fish is made for rapid swimming and high maneuverability~\cite{yu2018energy}.

The simple model of human throwing tasks
are studied by the principle of muscles contract in sequence~\cite{alexander1991optimum}.
%
And the sequence movement of the strong and weaker joint is also known as the Proximal-distal(P-D) sequence~\cite{herring1992effects,zatsiorsky2002kinetics}.
And then the timing accuracy in human over-arm throwing task is also studied~\cite{chowdhary1999timing}.
In Ref.~\cite{roach2013elastic}, the author collected the ball throwing data of 20 adults, 
and qualitatively gave the mechanism of human energy storage and release energy, and it also points out that only humans can regularly throw projectiles with high speed and accuracy compared to primates.
Besides, the suspended backpack for energy harvesting and reduced load impact is also studied~\cite{yang2021power}.
The suspended backpack is a typical coordinated motion in coupled system.


{{
In above, existing methods for trajectory generation in reflexive and rhythmic movements primarily rely on neural control mechanisms, with limited exploration of detailed human dynamics modeling. This limitation constrains their application in voluntary motion control, particularly in handling interactions between active and passive joints. To address this gap, we are the first to introduce a human dynamics-based voluntary motion mechanism into robot control, proposing and solving the corresponding optimal control model.

}}

\begin{figure}[t]
\captionsetup{font={footnotesize}}
\centering %
\includegraphics[width=0.36\textwidth]{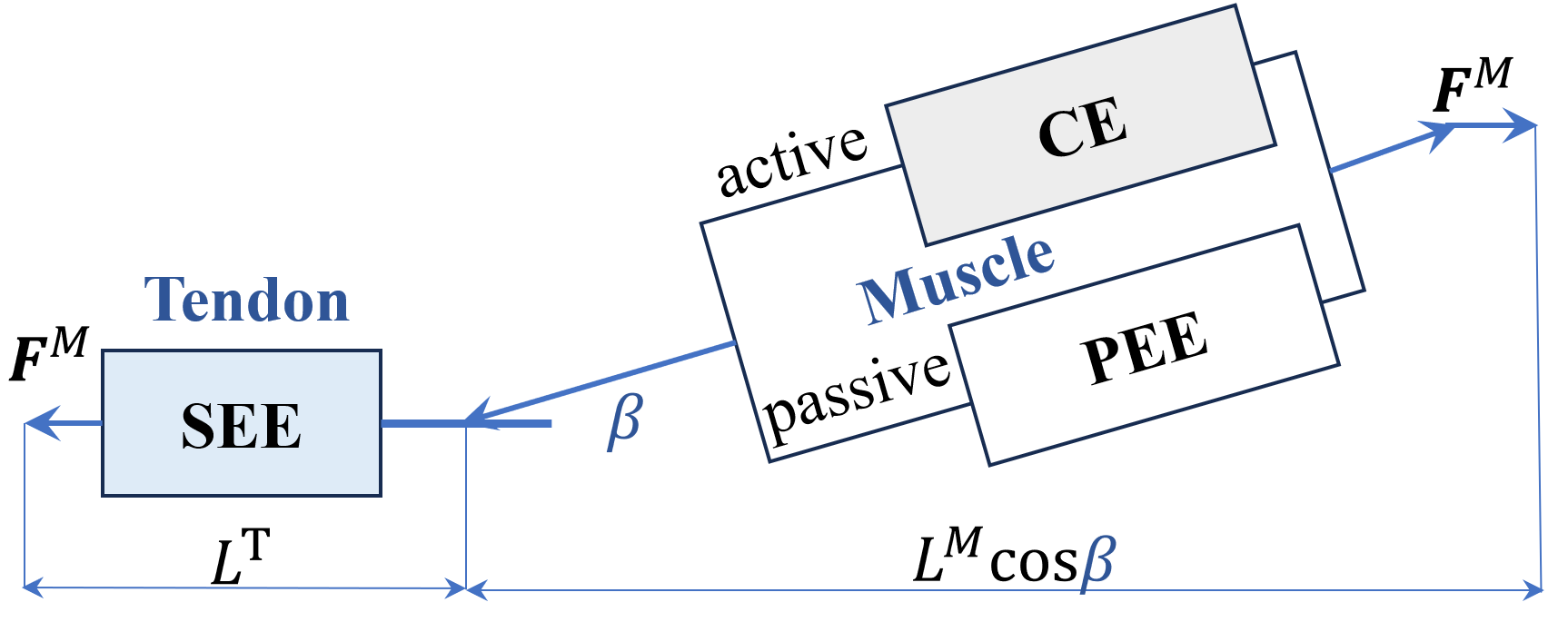}
\caption{The Hill-type musculotendon actuator. CE: Contractile element, PEE:
Parallel elastic element, SEE: Serial elastic element.
}
\vspace{-0.4cm}
\label{fig_musculotendon}
\end{figure}

\begin{figure}[t]
\captionsetup{font={footnotesize}}
\centering %
\includegraphics[width=0.5\textwidth]{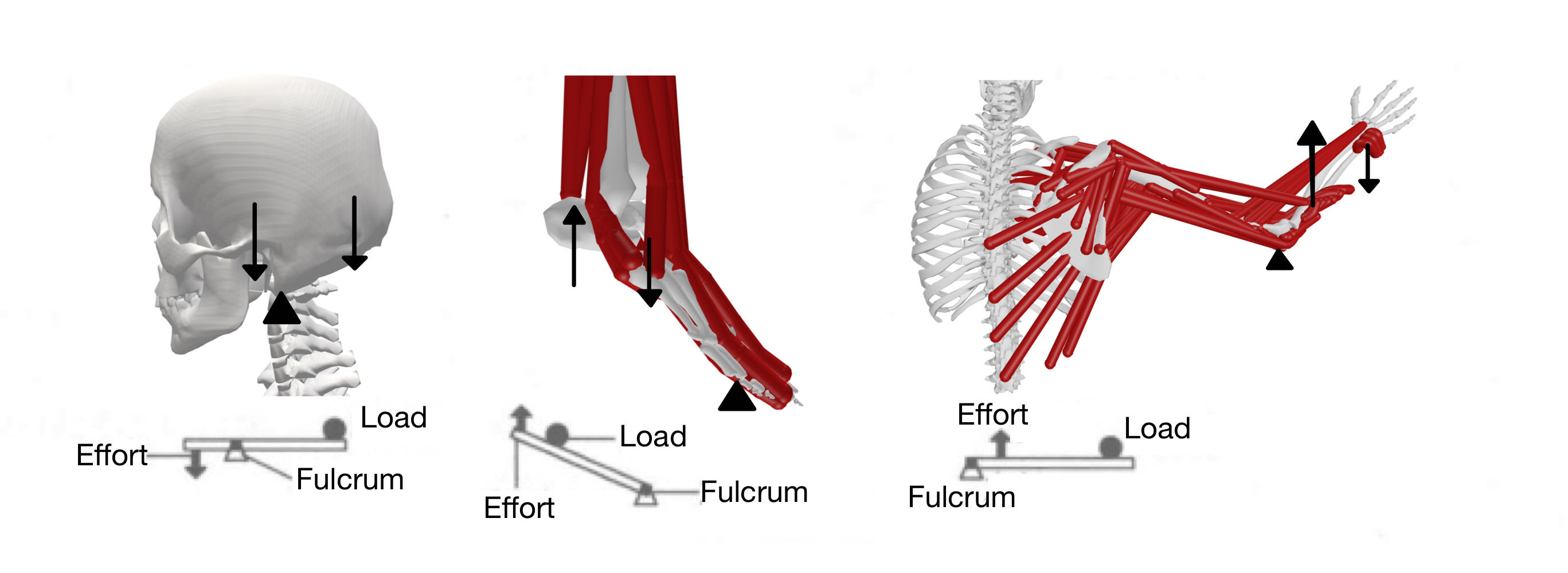}
\caption{The lever based mechanism of human movement.Balanced levers(left), labor-saving levers(middle), and labor-intensive levers(right)}
\vspace{-0.4cm}
\label{fig3}
\end{figure}

\section{The  Analysis of Human-like Motion}
\label{sec:III}

\subsection{Analysis of Human-like Motion Patterns}


We can identify two distinct actuator states from Fig.~\ref{fig_musculotendon}:

\begin{itemize}
    \item \textbf{Active state:} The actuator prioritizes achieving a specific position, speed, or external output force.
    \item \textbf{Passive state:} 
    The actuator follows external forces, which directly affect the target value.
\end{itemize}

Analyzing the Fig.~\ref{fig3}, 
we can extract key insights, particularly evident in throwing manipulations. The shoulder, elbow, and wrist-hand move in a sequential manner, with the laborious lever-based shoulder movement initiating the action and rapidly increasing speed. Subsequently, the delayed labor-saving lever-based wrist-hand
completes the action with explosive power at high speed. This sequence aligns with the proximal-distal (P-D) principle.
%
%
%

To represent this coordinated yet asynchronous movement, we introduce a response time matrix $\boldsymbol T$ for multiple joints, which also reflects the motion delay among different joints:
\begin{equation}
\boldsymbol T=
\begin{bmatrix}
t_{11} & t_{12} & \cdots & 0 & t_{1M}\\
t_{21} & 0 & \cdots & t_{2(M-1)} & t_{2M}\\
\vdots & \vdots  & \ddots   & \vdots & \vdots  \\
t_{N1} & 0 & \cdots & 0 & t_{NM}
\end{bmatrix}
\end{equation}
where $t_{ij}$ represents the time at which joint $i$ undergoes its $j$-th transition between active and passive modes, and $0$ indicates no mode change for that joint at that time. $N$ is the number of joints, and $M$ is the total number of time points considered.
For example, consider a throwing motion involving the shoulder (joint 1) and elbow (joint 2), the matrix could be:
\begin{equation}
\begin{bmatrix}
0.1 & 0.3 & 0.9 \\
0 & 0.2 & 0.9
\end{bmatrix}
\end{equation}
The shoulder starts active mode at \( t = 0.1 \, \text{s} \), while the elbow remains stationary (0 in the second row). The elbow begins its motion at \( t = 0.2 \, \text{s} \), showing a 0.1s delay. This delay between joint activations highlights motion delay in multi-joint movements. The matrix captures both the interaction between active and passive states across joints and the delays, as shown by the varying transition times and zero entries.

To systematically investigate human-like motion patterns, we formulate the fundamental
motor primitives as:
\begin{equation}
\label{eq:MPC_N}
\left\{
\begin{array}{lcl}
  \boldsymbol{\dot x} = \boldsymbol{A}\boldsymbol{x}+\boldsymbol{B}\boldsymbol{u}, \ 
  \left| \boldsymbol{u} \right| \leq  \boldsymbol{u}_{\text{max}} \\
  \boldsymbol{x}(0)=\boldsymbol{x}_0,
  \boldsymbol{x}(t_f) \in \boldsymbol{\chi}_{goal}
\end{array} 
\right.
\end{equation}
where $\boldsymbol{x}=[\theta; \dot{\theta}]$
, $\boldsymbol{A}=[0 \  1;0 \  0]$, $\boldsymbol{B}=[0 ;1]$
, $u=\ddot \theta$ denotes the 
control input, and $\boldsymbol{u}_\text{max}$ signifies the limit value of $\boldsymbol{u}$.

For each joint, we propose the following formulation for the cost function:
\begin{equation}
J(k_u, k_v, k_a) = k_u \cdot J_u - k_v \cdot J_v - k_a \cdot J_a
\end{equation}

Here, \( J_u = \int u^2 \, dt \) denotes the energy cost, \( J_v = \int \dot{\theta}^2 \, dt \) captures the velocity cost, and \( J_a = \int \ddot{\theta}^2 \, dt \) captures the acceleration cost. The weighting factors \( k_t \), \( k_u \), \( k_v \), and \( k_a \) balance the contributions of energy, velocity, and acceleration, respectively. In different modes, the weighting factors \(k_u\), \(k_v\), and \(k_a\) should be adjusted according to the optimization goals:

- \textbf{Active Mode}: To prioritize faster movement, increase the velocity and acceleration weights \(k_v\) and \(k_a\). The energy weight \(k_u\) can remain moderate.

- \textbf{Passive Mode}: Focus on energy optimization by increasing the energy weight \(k_u\), and reducing the velocity and acceleration weights \(k_v\) and \(k_a\).

- \textbf{Transition Mode}: Balance all weights \(k_u\), \(k_v\), and \(k_a\) to ensure smooth transitions while managing speed and energy.

We present three common scenarios to illustrate:

\begin{itemize}
    \item \textbf{Case I:} Distal joint active, proximal joint passive (e.g., Tai-Chi starting pose)
    \item \textbf{Case II:} Distal joint passive, proximal joint active (e.g., hitting a ball or waving)
    \item \textbf{Case III:} Alternating active and passive states between distal and proximal joints (e.g., throwing  an object)
\end{itemize}

These examples demonstrate how lever mechanisms, motion delay, and active/passive modes interact to produce complex, human-like movements.
This analysis informs the redesign of optimal control solutions
~\cite{de2012muscle}.

\subsection{Dynamics Analysis for Coordinated Movement}

Extending the principles of human-like motion patterns to multi-joint coordination, this approach applies to tasks like walking, robotic arm control, and sports techniques. The dynamics of these systems are described by coupled differential equations, where $\theta_i$ is the angular displacement of joint $i$, and $\tau_i$ is the applied torque. The motion equation for a system with $N$ joints is:

\begin{equation}
I_i \ddot{\theta}_i(t) = \tau_i(t, T) + \sum_{j=1}^{N} \mathcal{F}_{ij}(\theta_j, \dot{\theta}_j, \ddot{\theta}_j, t, T)
\end{equation}
where $I_i$ is the moment of inertia of joint $i$, and $\mathcal{F}_{ij}$ are the interaction forces and torques between joints.

To optimize coordination, we define an objective function \( J \) to capture performance goals, which can be expressed as:

\begin{equation}
J = \sum_{i=1}^{N} \alpha_i \cdot f_i(\tau_i(t), \theta_i(t), \dot{\theta}_i(t), \ddot{\theta}_i)
\end{equation}
where $f_i$ represents a function of the torque, angular displacement, and angular velocity for each joint, and $\alpha_i$ are weighting factors that balance the contributions of each joint to the overall performance. The goal is to find the torque functions $\tau_i$ and response times $t_{ij}$ that maximize $J$.

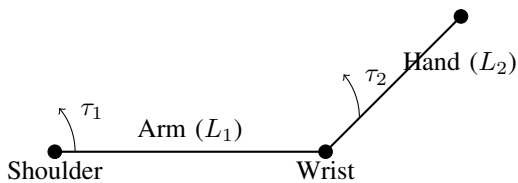
\begin{figure}[htbp]
\captionsetup{font={footnotesize}}
\centering
\begin{tikzpicture}[scale=0.9]  
  \draw[thick] (-1,0) -- (3,0) node[midway, above] {Arm ($L_1$)};
  \draw[fill=black] (-1,0) circle (0.1) node[below] {Shoulder};
  
  \draw[thick] (3,0) -- (5,2) node[midway, above right] {Hand ($L_2$)};
  \draw[fill=black] (3,0) circle (0.1) node[below] {Wrist};
  \draw[fill=black] (5,2) circle (0.1) node[right]{};
  
  \draw[->] (-0.7,0) arc[start angle=0, end angle=40, radius=1] node[midway, above right] {$\tau_1$};
  \draw[->] (3.5,0.5) arc[start angle=0, end angle=40, radius=1] node[midway, above right] {$\tau_2$};
\end{tikzpicture}
\caption{Coordinated motion analysis.
$L_1$ and $L_2$ are the length. $\tau_1$ and $\tau_2$ denote the shoulder (joint 1) and wrist (joint 2) torque respectively.}
\label{fig:four_object_system}
\end{figure}

With this framework, we apply it to a specific example: coordinated motion in a throwing task as shown in Fig \ref{fig:four_object_system}.

During the throwing motion, $\boldsymbol T$ is expressed as:
\begin{equation}
\boldsymbol T =
\begin{bmatrix}
t_{11} & t_{12} & t_{1M} \\
t_{22} & t_{23} & t_{2M}
\end{bmatrix}
\end{equation}
Here, $t_{11}$ is the time when the shoulder transitions from initial passive to active to start the throw, $t_{12}$ marks the shoulder's return to passive after applying torque, $t_{22}$ indicates when the wrist transitions from initial passive to active following the shoulder, and $t_{23}$ marks the wrist's return to passive after generating the necessary torque for release.

Furthermore,  $\tau_1$ and  $\tau_2$ can be described by:

\begin{equation}
I_1 \ddot{\theta}_1(t) = \tau_1(t, T) +  \mathcal{F}_{12}(t, T),\ \ 
I_2 \ddot{\theta}_2(t) = \tau_2(t, T)
\end{equation}

$\dot{\theta}_1$ and $\dot{\theta}_2$ determines the linear velocity at the hand's end: \(\mathbf{v}_f =  L_1 \dot{\theta}_1 \hat{\mathbf{r}}_1 + L_2 \dot{\theta}_2 \hat{\mathbf{r}}_2 \), where \( \hat{\mathbf{r}}_1 \) and \( \hat{\mathbf{r}}_2 \) are unit vectors perpendicular to the hand and arm, respectively, indicating the direction of their linear velocities due to rotation.

To maximize the distance of the thrown object, the objective function $J$ is defined as:
\begin{equation}
\begin{aligned}
\quad & J =  [v_f \cos(\theta_f) \cdot t_{\text{flight}}] \\
\text{subject to} \quad & \tau_{\min} \leq \tau_i(t, T) \leq \tau_{\max}, \quad i = 1, 2 \\
& \theta_{\min} \leq \theta_i(t) \leq \theta_{\max}, \quad i = 1, 2 \\
& P_{\min} \leq \tau_i(t, T)\dot{\theta}_{i}(t) \leq P_{\max}, \quad i = 1, 2 \\
\end{aligned}
\label{eq:11}
\end{equation}

where:
\begin{equation} 
\begin{aligned} 
v_f &= \sqrt{v_{f_x}^2 + v_{f_y}^2} \\
v_{f_x} &= L_2 \dot{\theta}_{2_f} \cos(\theta_{2_f}) + L_1 \dot{\theta}_{1_f} \cos(\theta_{1_f}) \\
v_{f_y} &= L_2 \dot{\theta}_{2_f} \sin(\theta_{2_f}) + L_1 \dot{\theta}_{1_f} \sin(\theta_{1_f}) \\
\theta_f &=  \underbrace{\int_{t_{22}}^{t_{2M}} \dot{\theta}_2(t) \, dt}_{\theta_{2_f}} + \underbrace{\int_{t_{11}}^{t_{1M}} \dot{\theta}_1(t) \, dt}_{\theta_{1_f}}\\
t_{\text{flight}} &= \frac{v_f \sin(\theta_f) + \sqrt{(v_f \sin(\theta_f))^2 + 2gh}}{g} \\
h &= L_2 \sin(\theta_f) + L_1  \sin(\theta_{1_f})
\end{aligned} 
\label{eq:12}
\end{equation}

It shows that \( J \) is a function of the joint angle variables, specifically \( J = f(\theta_1, \theta_2, \dot{\theta}_1, \dot{\theta}_2) \). Using the response time matrix \( T \), we can determine the state mode of each joint—whether it is active, passive, or in transition. Once the state mode is identified, the corresponding cost function is selected to optimize the joint torques:

\[
\min_{\tau_i(t)} J = \min_{\tau_i(t)} \int_{t_{i(j-1)}}^{t_{ij}} \left( k_u \cdot u^2 - k_v \cdot \dot{\theta}_i^2 - k_a \cdot \ddot{\theta}_i^2 \right) dt
\]

Once the optimal \( \tau_i(t) \) is obtained, the joint's angular acceleration \( \ddot{\theta}_i(t) \), velocity \( \dot{\theta}_i(t) = \int \ddot{\theta}_i(t) \, dt \), and position \( \theta_i(t) = \int \dot{\theta}_i(t) \, dt \) are computed using the dynamics equation:\( 
\ddot{\theta}_i(t) = \frac{1}{I_i} \left( \tau_i(t) + \sum_{j=1}^{N} \mathcal{F}_{ij}(\theta_j, \dot{\theta}_j, \ddot{\theta}_j, t) \right)
\). 

This hierarchical control structure ensures that all variables in equations~\ref{eq:11} and~\ref{eq:12}, such as torque and motion dynamics, are functions of \( T \), meaning \( J = f(T) \). This example illustrates how top-level control (identifying state mode with \( T \)) and bottom-level control (optimizing torque and motion) work together to efficiently manage joint movement.

\subsection{Compliant Motion of Passive Joint}
The dynamic equation of the robot can be expressed as:
\begin{equation}
\boldsymbol M_a \ddot{\boldsymbol \theta}_a +\boldsymbol M_{ap} \ddot{\boldsymbol \theta}_{p} + {\boldsymbol C}_a + {\boldsymbol G}_a 
= \boldsymbol \tau_a
+ \boldsymbol \tau_{ac}
\end{equation}
\begin{equation}
\boldsymbol M_p \ddot{\boldsymbol \theta}_p +\boldsymbol M_{pa} \ddot{\boldsymbol \theta}_{a} + {\boldsymbol C}_p + {\boldsymbol G}_p 
= \boldsymbol \tau_p
+
\boldsymbol \tau_{pc}
\end{equation}
where subscripts '$a$' and '$p$' denote the active and passive joints respectively, subscripts '$ap$' and '$pa$' denote the coupling term.
$\boldsymbol M$, $\boldsymbol C$, and $\boldsymbol G$ are the mass matrix, Coriolis and centrifugal force matrix, gravitational torque vector respectively.
\(\mathbf{ \boldsymbol \tau} = \begin{bmatrix} {\boldsymbol \tau}^T_a & {\boldsymbol \tau}^T_p \end{bmatrix}^T\) is the control torque vector.
\(\mathbf{ \boldsymbol \tau}_c = \begin{bmatrix} {\boldsymbol \tau}^T_{ac} & {\boldsymbol \tau}^T_{pc} \end{bmatrix}^T\) is the contact torque vector.
\(\mathbf{ \boldsymbol \theta} = \begin{bmatrix} {\boldsymbol \theta}^T_a & {\boldsymbol \theta}^T_p \end{bmatrix}^T\) is the system state vector.

The compliant motion mechanism, modeled as inertia-damping-spring system, effectively buffers the forces between active and passive joints.
\begin{equation}
\boldsymbol B_p {\Delta \boldsymbol { \ddot \theta}}_p +
{\boldsymbol D_p}{\Delta \boldsymbol { \dot \theta}}_p 
+
{\boldsymbol K_p
{\Delta \boldsymbol { \theta}}_p 
=
\boldsymbol { \tau}}_{pr}
\end{equation}
where $\boldsymbol B_{p}$, $\boldsymbol D_{p}$ and $\boldsymbol K_{p}$ are the human-like inertia, damping and stiffness for passive joint respectively. 
${\Delta \boldsymbol {  \theta}}_p={\boldsymbol {\theta}}_{p,\text{ref}}-{\boldsymbol {  \theta}}_p$
Subscript '$\text{ref}$' denotes the desired value.
The actual torque $\boldsymbol {\tau}_{pr}$ can be measured using a joint torque sensor or calculated by inverse dynamics.
The  compliant behavior of passive joint will be affected by $\boldsymbol {\tau}_{pr}$.

\section{The Optimal Control Solution}
\label{sec:IV}

We use the mechanism of human motion to re-customize the following optimal control solution. Let $ \boldsymbol{\dot x}=[\boldsymbol\theta^T \ {\dot {\boldsymbol \theta^T} }]^T$.
%
\begin{equation}
\label{eq:MPC_N}
\left\{
\begin{array}{lcl}
\min J= \Phi(\boldsymbol x(t_f)) + \int_{o}^{t_f}L(\boldsymbol x,t)  dt\vspace{1ex}
 \\
{\rm s.t}:
\  \boldsymbol{\dot x}=f(\boldsymbol{x},\boldsymbol{u})   \vspace{1ex},
\\
\ \ \ \ \ \ \ \boldsymbol{u} \in \boldsymbol{U}
\\
\ \ \ \ \ \ \  \boldsymbol{g}(\boldsymbol{x},\boldsymbol{u}) = 0  
\\
\ \ \ \ \ \ \  \boldsymbol{h}(\boldsymbol{x},\boldsymbol{u}) \leq 0
\\
\ \ \ \ \ \ \  \boldsymbol{x}(0)=\boldsymbol{x}_0,\boldsymbol{x}(t_f) \in \boldsymbol{\chi}_{goal} 
\end{array} 
\right.
\end{equation}
where $L(\boldsymbol x,t)$, and $\Phi(\boldsymbol x(t_f))$ denote the running  and the terminal cost. $\boldsymbol{g}(\boldsymbol{x},\boldsymbol{u})$, $\boldsymbol{h}(\boldsymbol{x},\boldsymbol{u})$ are the equality constraints and inequality constraints respectively. 
%
%

\subsection{The Human-inspired Motion Planning Algorithm}
%
%
The velocity and acceleration of each joint subsystem are decoupled from each other in terms of kinematics, but from a dynamic point of view (such as torque constraints, power constraints), each subsystem is coupled to each other. Therefore, it is necessary to reformulate the whole system.

\subsubsection{Constrained Optimal Control for Human-like Motion}
The whole system should satisfy the total energy constraints as:
\begin{equation}
\sum_{i=1}^{N} |\tau_i \dot \theta_i| \leq P_{\text{max}}
\label{eq:energy_constraint}
\end{equation}
where $P_{\text{max}}$ is the maximum system power. Let $\tau_{i,\text{min}}$ and $\tau_{i,\text{max}}$ be the limit torque value. Each joint needs to satisfy:
\begin{equation}
|\tau_i| \leq \tau_{i,\text{max}}
\label{eq:torque_constraint}
\end{equation}
Additionally, each joint state $\theta_i$ must satisfy:
\begin{equation}
\theta_{i,\text{min}} \leq \theta_i \leq \theta_{i,\text{max}}
\label{eq:joint_state_constraint}
\end{equation}
where $\theta_{i,\text{min}}$ and $\theta_{i,\text{max}}$ are the minimum and maximum allowable joint states for the $i$-th joint, respectively.
Furthermore, the system dynamics equation is:
\begin{equation}
\dot{\boldsymbol x}(t) = f(\boldsymbol x(t), \boldsymbol u(t, T), t), \quad t_0 \text{ and } \boldsymbol x(t_0) \text{ are given.}
\label{eq:dynamics_constraint}
\end{equation}
For human motion planning, the cost function $J$ can be formulated as a combination of different objectives. Based on (\ref{eq:MPC_N}), a general form of the cost function can be:
\begin{equation}
J = \int_{t_0}^{t_f} L(\boldsymbol x(t), \boldsymbol u(t, T), t) dt + \Phi(\boldsymbol x(t_f), t_f)
\label{eq:cost_function}
\end{equation}

The optimal motion trajectory is then obtained by minimizing this cost function.

\subsubsection{Augmented Hamiltonian for Constrained Optimal Control}

We define an augmented costate vector $\Lambda(t)$ that includes both $\lambda(t)$ and $\mu$:
\begin{equation}
\Lambda(t) = 
\begin{bmatrix}
\lambda(t) & \mu
\end{bmatrix}^T
=
\begin{bmatrix}
\lambda_1(t) \cdots  \lambda_n(t) & \mu_1 \cdots \mu_M
\end{bmatrix}^T
\end{equation}
where,
$\lambda(t) \in \mathbb{R}^n$ is the original lagrangian multiplier.
$\mu \in \mathbb{R}^M$ is the vector of multipliers for inequality constraints.
$n$ is the dimension of the state space.
$M$ is the number of inequality constraints.

Now, we can rewrite the Hamiltonian:
\begin{equation}
H(\boldsymbol x, \boldsymbol u, \Lambda, t, T) = L(\boldsymbol x, \boldsymbol u, t) + \Lambda^T(t)\mathbf{G}(\boldsymbol x, u, t, T)
\end{equation}

The vector $\mathbf{G}$ includes all constraints: $\mathbf{G}(\boldsymbol x, \boldsymbol u, t, T) = \begin{bmatrix}
f(\boldsymbol x, \boldsymbol u, t, T) \\
g(\boldsymbol x, \boldsymbol u)
\end{bmatrix}$, 
where $f(\boldsymbol x, \boldsymbol u, t, T)$ represents the system dynamics, and $g(\boldsymbol x, \boldsymbol u)$ is the inequality constraint vector. Similar to Eq.~\ref{eq:12}, for a given $T$, the control input $u_i(t, T)$ can be determined at any time, allowing us to compute $x_i(t)$. Consequently, $H$ can be evaluated at any time based on $T$.

The necessary conditions for optimality are:
\begin{align}
\dot{\boldsymbol x} &= f(\boldsymbol x, \boldsymbol u, t, T) \\
\dot{\Lambda}(t) &= 
\begin{bmatrix}
\dot{\lambda}(t) \\
\dot{\mu}(t)
\end{bmatrix}
= 
\begin{bmatrix}
-\frac{\partial L}{\partial \boldsymbol x} - \frac{\partial f}{\partial \boldsymbol x} \lambda(t) \\
\dot{\mu}(t)
\end{bmatrix}\\
{\partial H}/{\partial u} &= 0
\end{align}

The value variance of $\mu(t)$ depends on complementary slackness conditions:
\begin{equation}
\begin{cases}
g_j(\boldsymbol x, \boldsymbol u) < 0 \implies \mu_j = 0 \\
g_j(\boldsymbol x, \boldsymbol u) = 0 \implies \mu_j \geq 0 \\
\end{cases}
\quad \forall j = 1, ..., M
\end{equation}

The sufficient condition for optimality is: 
\begin{equation}
    \partial_u^2 H({\boldsymbol x}^*, \boldsymbol u^*, \Lambda^*, t, T^*) > 0
\end{equation} where $(\cdot)^*$ represent optimum.

We use these conditions to solve for the optimal control input as a function of \( T \). When an analytical solution is not feasible, numerical solvers can be applied, often by solving a two-point boundary value problem (BVP) using collocation.

The feasibility region $\mathcal{F}$ is defined as the intersection of constraint sets: $\mathcal{F} = \bigcap_{i=1}^{m} \{(\boldsymbol x, \boldsymbol u) \mid g_i(\boldsymbol x, \boldsymbol u) \leq 0\}$. As $m$ increases, the volume of $\mathcal{F}$ decreases, potentially leading to an empty set. A smaller region makes finding the optimal solution harder, often near the boundary, complicating algorithmic convergence. This is quantified by the violation measure: $V(\boldsymbol x,\boldsymbol u) = \sum_{i=1}^{m} \max(0, g_i(\boldsymbol x,\boldsymbol u))$. The probability of a feasible solution decreases exponentially with $m$: $P(\text{feasible}) \propto e^{-\alpha m}$, where $\alpha$ is a problem-specific constant.

Constraints vary in criticality. Critical constraints, like joint angle limits preventing mechanical damage, must be guaranteed. Less critical constraints, such as joint acceleration limits for smooth operation, can be slightly violated. Their satisfaction is encouraged rather than strictly enforced. Therefore, we reformulate the augmented Hamiltonian by incorporating penalty functions for less critical constraints. This approach reformulates the augmented Hamiltonian as:
\begin{equation}
H_{a} = H + \sum_{j=1}^{M} P_j(g_j(\boldsymbol x,\boldsymbol u)), \quad \text{with } g_j(\boldsymbol x,\boldsymbol u) \leq 0
\end{equation}
where $P_j(\cdot)$ is the penalty function for the $j$-th constraint. For critical constraints, we use $P_j(g_j(\boldsymbol x, \boldsymbol u)) = \mu_j g_j(\boldsymbol x, \boldsymbol u)$. For less critical constraints, a quadratic penalty is applied:
\begin{equation}
P_j(g_j(\boldsymbol x, \boldsymbol u)) = 
\begin{cases}
    0 & \text{if } g_j(\boldsymbol x, \boldsymbol u) \leq 0 \\
    k_j (g_j(\boldsymbol x, \boldsymbol u))^2 & \text{if } g_j(\boldsymbol x, \boldsymbol u) > 0
\end{cases}
\end{equation}
where $k_j$ is the penalty parameter for the $j$-th constraint. By prioritizing critical constraints and tolerating small violations of less critical constraints, the system expands the feasibility region $\mathcal{F}$, making it easier to find feasible solutions under complex constraint sets.


\subsubsection{Optimal Control for Joint Motion}
Our algorithm solves the motion of each joint under kinematic constraints and parameterizes the motion time of both active and passive joints. This transforms the entire robot system into a time-parameterized optimal control problem. 
The algorithm is as the  following \textbf{Algorithm 1}.

\begin{algorithm}[htbp] 
    \footnotesize
    \caption{Human Inspired Optimal Control}\label{alg:human_inspired_optimal_control}
    \begin{algorithmic}
    \STATE \textbf{Given:} $\boldsymbol{x}_0$, $\boldsymbol{x}_g$, performance index $L$, constraints $\{g_j(x,u)\}_{j=1}^M$
    \STATE \textbf{Initialize:} $\boldsymbol{q}_0 \gets \boldsymbol{x}_0$, $\boldsymbol{q}_g \gets \boldsymbol{x}_g$, Initialize response time matrix $T$
    \STATE Classify constraints into critical and less critical sets
    \STATE Define penalty functions $P_j$ for each constraint
    \STATE
    \STATE \textbf{Function} \textsc{EvaluateMotion}$(T)$:
    \STATE \hspace{0.4cm} Initialize empty sets $\Theta$, $\dot{\Theta}$, $\ddot{\Theta}$, $\mathcal{T}$
    \STATE \hspace{0.4cm} \textbf{for each} joint $k$ \textbf{from} $N$ \textbf{to} 1 \textbf{do}
    \STATE \hspace{0.8cm} Define joint state based on $T$
    \STATE \hspace{0.8cm} Formulate cost function for each state interval
    \STATE \hspace{0.8cm} Solve for $\tau_k(t, T)$ based on compliant motion dynamics 
    \STATE \hspace{0.8cm} Add $\tau_k(t, T)$ to $\mathcal{T}$
    \STATE \hspace{0.8cm} Compute $\ddot{\theta}_k(t, T)$, $\dot{\theta}_k(t, T)$, $\theta_k(t, T)$
    \STATE \hspace{0.8cm} Add $\ddot{\theta}_k(t, T)$ to $\ddot{\Theta}$, $\dot{\theta}_k(t, T)$ to $\dot{\Theta}$, $\theta_k(t, T)$ to $\Theta$
    \STATE \hspace{0.4cm} \textbf{end for}
    \STATE \hspace{0.4cm} Compute base Hamiltonian $H$ using $\{\Theta, \dot{\Theta}, \ddot{\Theta}, \mathcal{T}\}$
    \STATE \hspace{0.4cm} Compute augmented Hamiltonian:
    \STATE \hspace{0.8cm} $H_a(T) = H + \sum_{j=1}^M P_j(g_j(x,u))$
    \STATE \hspace{0.4cm} \textbf{return} $\{H_a(T), \Theta, \dot{\Theta}, \ddot{\Theta}, \mathcal{T}\}$
    \STATE
    \STATE $T^* \gets \arg\underset{T}{\max} H_a(T)$ \textbf{from} \textsc{EvaluateMotion}$(T)$
    \STATE
    \STATE \textbf{for each} joint $k$ \textbf{from} 1 \textbf{to} $N$ \textbf{do}
    \STATE \hspace{0.4cm} Generate optimal trajectory $q_k(t)$, $\dot{q}_k(t)$, $\ddot{q}_k(t)$ using $T^*$
    \STATE \hspace{0.4cm} Compute optimal torque profile $\tau_k(t)$ using $T^*$
    \STATE \textbf{end for}
    \STATE
    \STATE Apply trajectory and force tracking control (e.g., impedance, admittance control, or their combination with whole body dynamic control)
    \end{algorithmic}
\end{algorithm}

\subsection{Algorithm Extension:Precision-Compliance Trade-off}

To achieve an optimal balance between precise control and system compliance, it is essential to enhance both operational accuracy and adaptability to external environments and uncertainties. This adaptability is crucial for improving safety, stability, and efficiency in executing complex tasks.

The system's compliance is determined by joint inertia, damping, and stiffness, and the compliance mechanism of the passive joint is described by its response to external torques (see Eq.15). Compliance is inversely related to system stiffness and damping:
\begin{equation}
    \gamma_{\theta} \propto \frac{1}{K_p + D_p}, \quad \gamma_{\dot{\theta}} \propto \frac{1}{K_p + D_p}
\end{equation}

Increasing stiffness \(K_p\) and damping \(D_p\) reduces compliance, limiting system's adaptability to external disturbances.

In contrast, precise control aims to minimize the position and velocity errors of both active and passive joints:
\begin{equation}
e_{\theta_p} = \theta_{p,\text{ref}} - \theta_p, \quad e_{\dot{\theta}_p} = \dot{\theta}_{p,\text{ref}} - \dot{\theta}_p
\end{equation}

Reducing these errors requires increasing stiffness and damping:
\begin{equation}
e_{\theta_p} \propto \frac{1}{K_p}, \quad e_{\dot{\theta}_p} \propto \frac{1}{D_p}
\end{equation}

Thus, increasing stiffness and damping enhances control precision but reduces compliance. The balance between them should be dynamically adjusted according to task requirements and environment by tuning stiffness and damping for optimal control.

\section{Experiments}
\label{sec:V}

In this paper, we prepare two types of experiments: dual UR16e arms based  throwing manipulation, and motion control of a certain type of multi-DOF humanoid robot.

\subsection{Experiments on a Collaborative Robotic Arm}

As UR16e arms are limited by  the total power constraints ($P_{\text{max}}$=350 W) and joint velocity limit ($\dot \theta_{\text{max}}$=3.14 rad/s), this makes them easy to stop suddenly when executing dynamic tasks.
Therefore, it is necessary to calculate the  optimal trajectory of the robot, 
and we select the terminal time  $t_f$ and  $\tau^2$ as the cost function.
The robot arm uses position control and admittance control to implement the algorithm in this paper.


The proximal joints of UR16e are joint 1,2,and 3. And the  distal joints are the 4,5, and 6. 
This experiment is conducted in a vertical gravity field and only uses joints 2~(shoulder), 3~(elbow), and 4~(wrist) of the robot arm to throw object.
%

\begin{figure}[t]
\captionsetup{font={footnotesize}}
\centering

\subfigure[Proximal joints for throwing.]{
\begin{minipage}[t]{0.46\linewidth}
\centering
\includegraphics[width=1.6in]{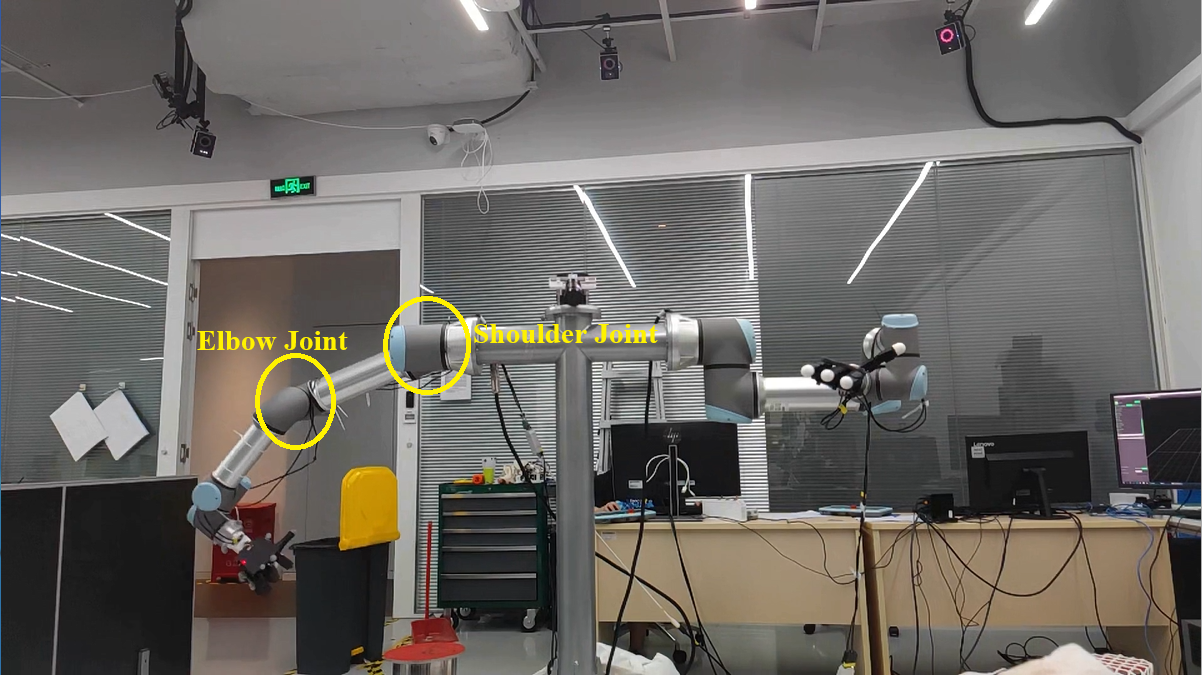}
\end{minipage}%
}%
\subfigure[Distal joints for throwing.]{
\begin{minipage}[t]{0.46\linewidth}
\centering
\includegraphics[width=1.6in]{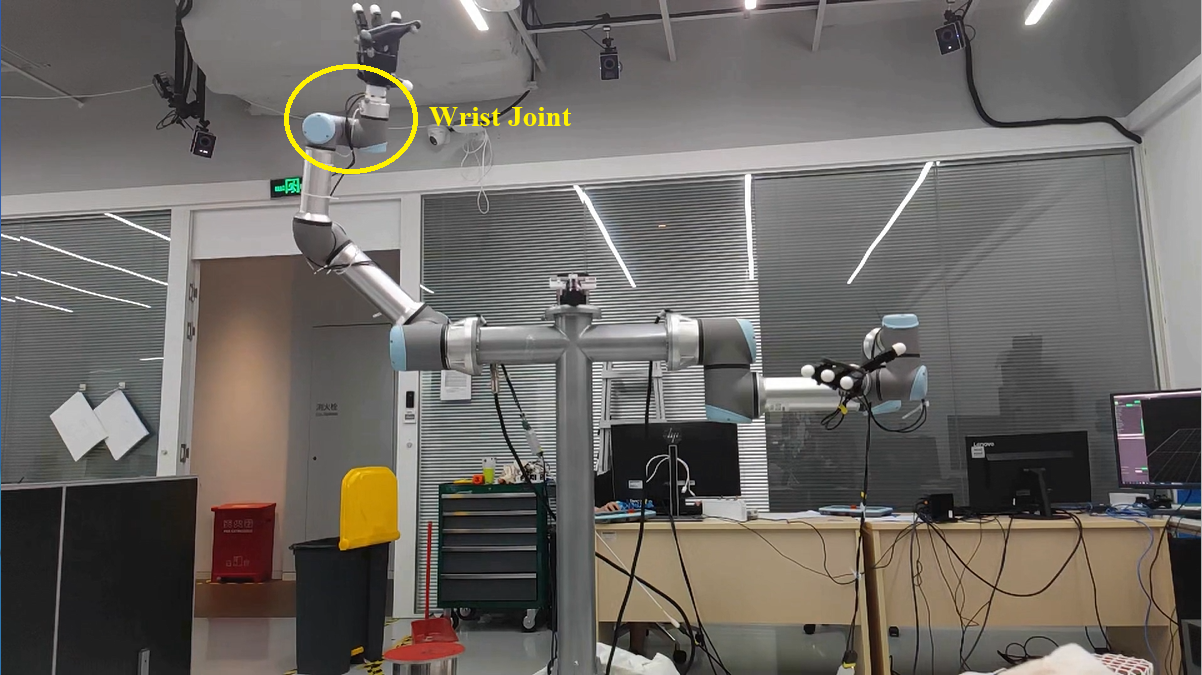}
\end{minipage}%
}%


\centering
\caption{The throwing motion of shoulder-elbow-wrist type robot. }
%
\end{figure}

\begin{figure}[t]
\captionsetup{font={footnotesize}}
\centering %
\includegraphics[width=0.45\textwidth]{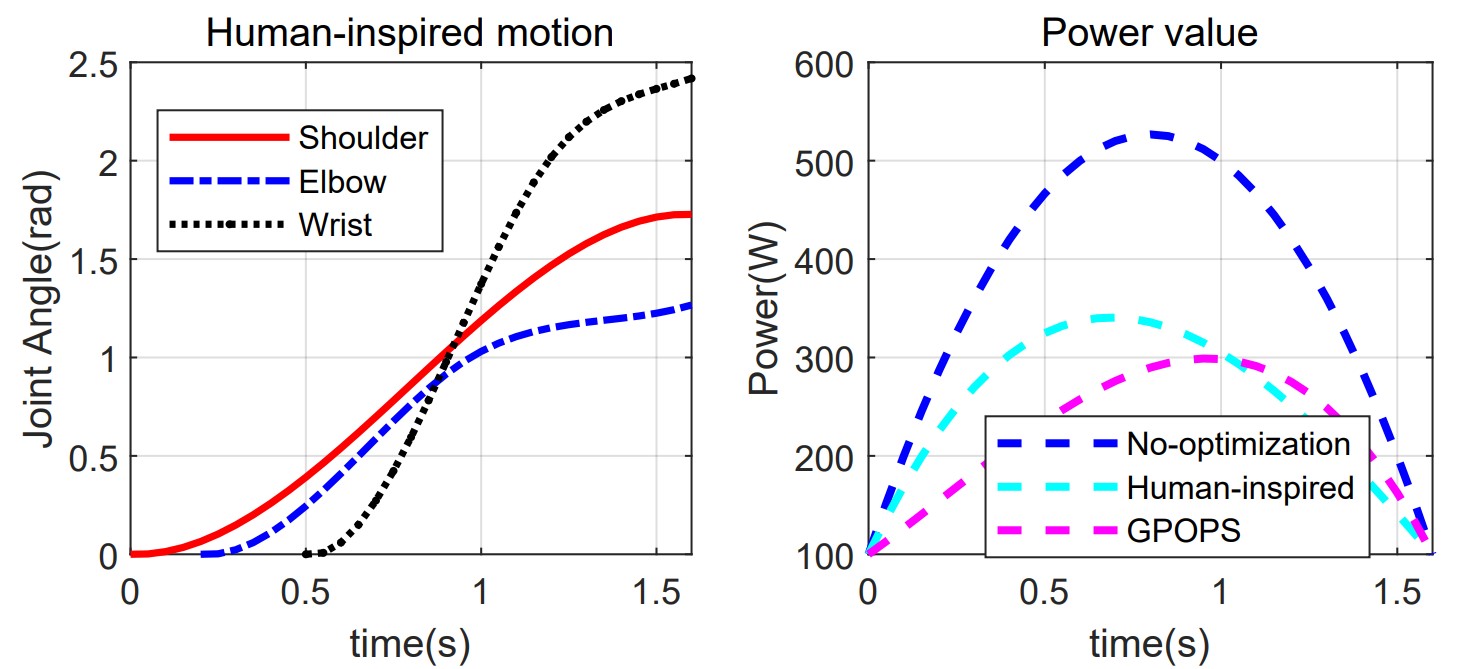}
\caption{The curves of the dynamic throwing manipulation.}
\vspace{-0.4cm}
\label{fig2}
\end{figure}

A screenshot of the experiment is shown in Fig.5. 
As can be seen from the Fig.6,
the results follow the P-D principle.
The proximal joints (shoulder and elbow) of UR16e move sequentially to obtain a larger linear velocity. The distal joint~(wrist) moves last to throw the object to minimize the impact of this action on the acceleration of the proximal joints.
When the distal joint completes the throwing task, multiple joints of the robotic arm are in a passive state to buffer the impact of dynamic motion on the joints.
%

The response time matrix for the right arm (throwing the object) is derived by nonlinear optimization. 
\begin{equation}
\boldsymbol T=
\begin{bmatrix}
\epsilon & 0  & 1.5 & 1.6\\
0 & 0.2 &  1.2 & 1.6\\
0 & 0.5  & 1.4 & 1.6
\end{bmatrix}
\end{equation}
where $\epsilon$ is a very small value.

In Fig.5. we can see the motion trajectory  of each joint switching from an active joint to a passive joint.
And the shoulder joint is active in $(\epsilon-1.5)$s and passive in $(1.5-1.6)$s. The elbow joint is active in $(0.2-1.2)$s, and passive in $(1.2-1.6)$s. The wrist joint is active in  $(0.5-1.4)$s, and passive in $(1.4-1.6)$s.
%
%
In Fig.6, we use the total power as the measurement indicator. Under the same boundary conditions, the algorithm proposed in this paper can achieve about 80\% of the offline optimal control software package GPOPS~\cite{patterson2014gpops}, and is significantly better than the result without optimization.

\subsection{Experiments on a Novel Humanoid Robot}

We chose a humanoid robot that is being developed in our laboratory. The total weight is about 80kg. This robot has a large load capacity. So an efficient motion trajectory generation algorithm is crucial in achieving locomotion-manipulation tasks.
Besides, the driving torque of hip and leg joints are limited by 
200Nm and 800N. 
The large inertia of the upper body requires the robot's leg-hip linkage to form a composite optimal motion trajectory during the configuration adjustment of the lower body, otherwise the robot's leg-hip will be subject to a large load.
And we select  $t_f$ and  $\tau^2$ as the cost function. 
The robot adopts impedance control and whole-body dynamics controller to implement the proposed algorithm.

A screenshot of the experiment is shown in Fig.7. 
In the standing stage, since the robot can be regarded as a planar robot at this stage, we only take the three joints leg-hip-waist (regarded as joint 1,2,3 respectively) as  example to illustrate.
Furthermore, $\boldsymbol T$ is derived as:
\begin{equation}
\boldsymbol T=
\begin{bmatrix}
0.6 & 0  & 3 \\
\epsilon & 0 &  3\\
0 & 0  & 3 
\end{bmatrix}
\end{equation}

So we can see that the leg-hip joint is active  in $(\epsilon-3)$s. In order to coordinate with the movement of the hip joint, the leg joints have a 0.6s delay.
The waist joint is passive in $(0-3)$s, and then it turns into active state.
From Fig.8, during the robot's standing phase, hip joint moves first, followed by leg joint. The corresponding optimal motion trajectory can reduce the peak power by about 20\%.
During the whole process, the waist joint needs to effectively cooperate with the movement of other joints. Therefore, the waist joint will switch from a passive joint to an active joint, and in this process, its reference position remains unchanged. The waist joint will have a certain buffering movement under impact.
%
%

\begin{figure}[t]
\captionsetup{font={footnotesize}}
\centering

\subfigure[]{
\begin{minipage}[t]{0.3\linewidth}
\centering
\includegraphics[width=0.98in]{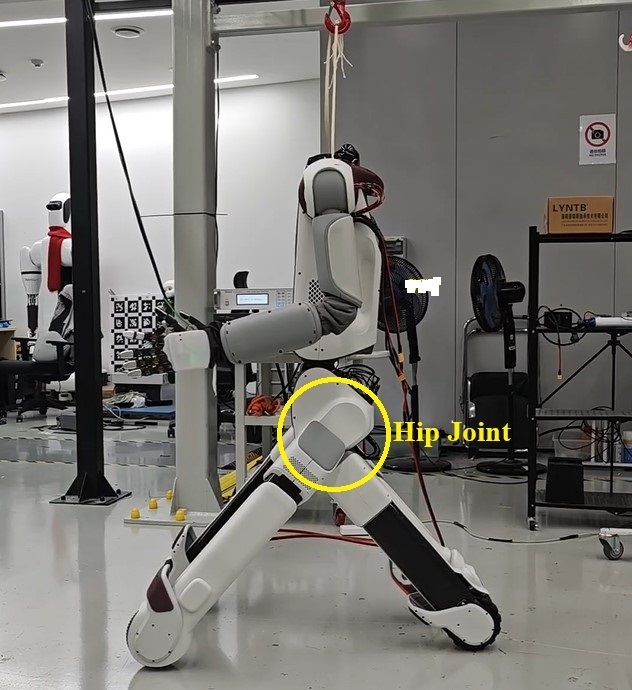}
\end{minipage}%
}%
\subfigure[]{
\begin{minipage}[t]{0.3\linewidth}
\centering
\includegraphics[width=1.0in]{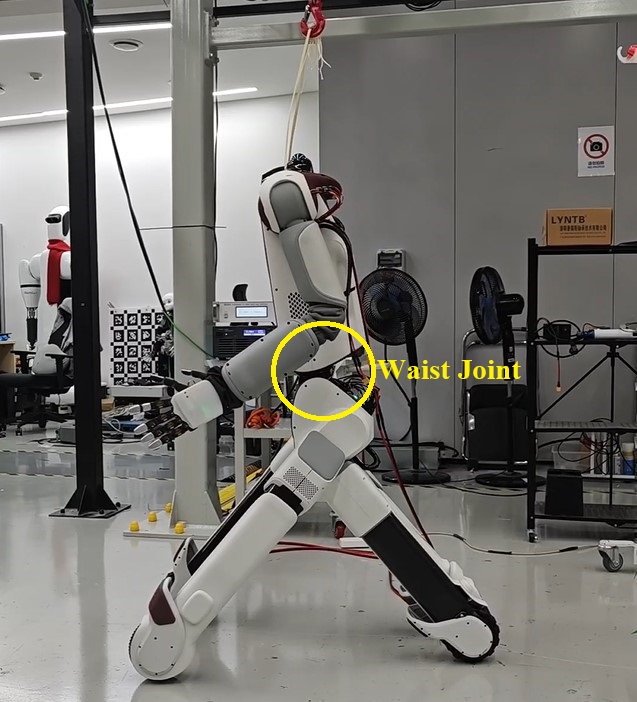}
\end{minipage}%
}%
\subfigure[]{
\begin{minipage}[t]{0.3\linewidth}
\centering
\includegraphics[width=1.0in]{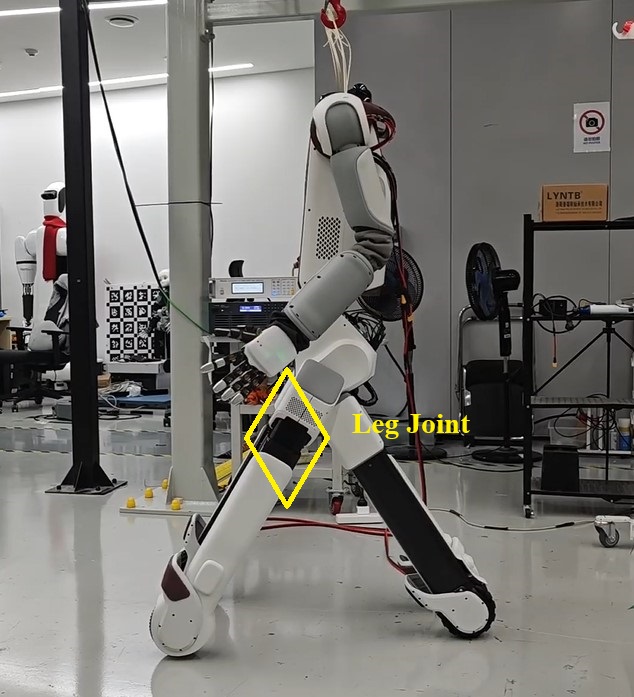}
\end{minipage}%
}%

\centering
\caption{Coordinated motion of leg-hip-waist type robot. (a) Hip joint moves firstly. (b) Waist joint is passive joint. (c) Leg joint moves after hip joint.}




\end{figure}
\begin{figure}[t]
\captionsetup{font={footnotesize}}
\centering %
\includegraphics[width=0.4\textwidth]{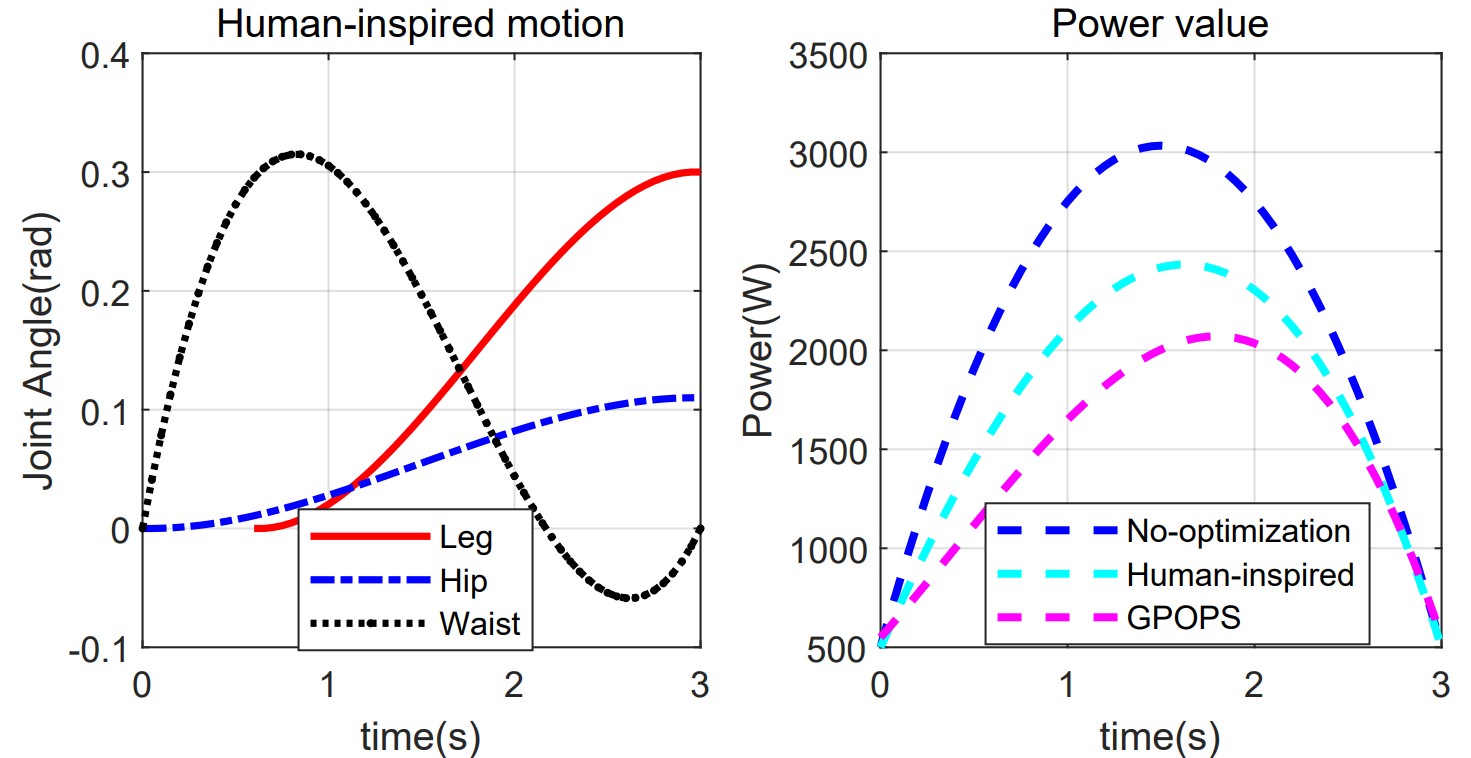}
\caption{The curves of the dynamic locomotion.}
\vspace{-0.2cm}
\label{fig2}
\end{figure}

\section{Conclusions}
\label{sec:VI}
This paper presents an optimal control framework based on compliant control dynamics, optimizing robotic arm motion by solving the joint response time matrix. This method significantly enhances the robot's compliance under external disturbances and achieves precise trajectory generation and control in complex tasks. Experimental results demonstrate that this framework notably improves the flexibility and stability of the robotic arm, ensuring its reliability in human-robot interactions.

\bibliographystyle{IEEEtran}
\normalem
\bibliography{ref}

\end{document}